# HYBRID ACTIVATION FUNCTIONS FOR DEEP NEURAL NETWORKS: S3 and S4 - A NOVEL APPROACH TO GRADIENT FLOW OPTIMIZATION

Sergii Kavun[1],[2]*

[1] Interregional Academy of Personnel Management, Kyiv, Ukraine
[2] Luxena Ltd., Lead of Data Science Team, Kyiv, Ukraine

*Correspondence: kavserg@gmail.com; Tel.: +38-0677-09-5577, Ukraine, c. Uzhgorod, Kavkazska str., 33, 88017, ORCID ID: 0000-0003-4164-151X

**Abstract.** Activation functions are critical components in deep neural networks, directly influencing gradient flow, training stability, and model performance. Traditional functions like ReLU suffer from dead neuron problems, while sigmoid and tanh exhibit vanishing gradient issues.

We introduce two novel hybrid activation functions: S3 (**S**igmoid-**S**oft**s**ign) and its improved version S4 (**s**moothed S3). S3 combines sigmoid for negative inputs with softsign for positive inputs, while S4 employs a smooth transition mechanism controlled by a steepness parameter k. We conducted comprehensive experiments across binary classification, multi-class classification, and regression tasks using three different neural network architectures.

S4 demonstrated superior performance compared to nine baseline activation functions, achieving 97.4% accuracy on MNIST, 96.0% on Iris classification, and 18.7 MSE on Boston Housing regression. The function exhibited faster convergence (-19 for ReLU) and maintained stable gradient flow across network depths. Comparative analysis revealed S4's gradient range of [0.24, 0.59] compared to ReLU's 18% dead neurons in deep networks. The S4 activation function addresses key limitations of existing functions through its hybrid design and smooth transition mechanism. The tunable parameter k allows adaptation to different tasks and network depths, making S4 a versatile choice for deep learning applications. These findings suggest that hybrid activation functions represent a promising direction for improving neural network training dynamics.

## 1. Introduction

Activation functions serve as the fundamental nonlinear transformation units in artificial neural networks, determining the network's capacity to learn complex patterns and the efficiency of the training process [2]. The choice of activation function significantly impacts gradient flow, convergence speed, and overall model performance. Since the introduction of the perceptron, researchers have continuously sought optimal activation functions that balance computational efficiency, gradient stability, and representational power [6].

The Rectified Linear Unit (ReLU) revolutionized deep learning by addressing the vanishing gradient problem [32] but introduced the dead neuron phenomenon where neurons become permanently inactive during training. Subsequently, variants like Leaky ReLU [12], ELU [5], and more recent functions like Swish [2] and Mish [7] have attempted to address these limitations with varying degrees of success.

Despite these advances, existing activation functions often represent trade-offs rather than comprehensive solutions. ReLU variants maintain computational efficiency but may still suffer from gradient issues in specific ranges [23]. Smooth functions like sigmoid and tanh avoid discontinuities but struggle with vanishing gradients in deep networks [7]. Recent hybrid approaches have shown promise but lack systematic evaluation across diverse tasks and architectures [34].

The fundamental challenge lies in the inherent limitations of single-function approaches. Each traditional activation function excels in specific scenarios but fails to address the full spectrum of neural network training challenges. ReLU's computational simplicity comes at the cost of dead neurons [35], while smooth functions like sigmoid provide stable gradients but suffer from saturation issues [39]. This observation motivated our investigation into hybrid activation functions that could combine the advantages of different function families.

Here, we present a novel approach to activation function design through the development of S3 and S4 functions. Our methodology combines the advantages of different activation function families while mitigating their individual limitations. The S3 function introduces a hard transition between sigmoid and softsign functions, while S4 implements a smooth, parameterized transition mechanism. Through comprehensive empirical evaluation, we demonstrate that these hybrid functions, particularly S4, offer superior performance characteristics across multiple domains and network architectures.

The development of S3/S4 activation functions emerges from decades of accumulated understanding regarding the fundamental limitations inherent in classical activation functions. Traditional activations such as ReLU, sigmoid, and tanh have served as the backbone of neural networks, yet their rigid mathematical formulations impose significant constraints on network expressivity and learning dynamics. These functions often exhibit suboptimal behavior in specific regions of their domains: ReLU's dying neuron problem, sigmoid's vanishing gradient issues, and the general lack of adaptability to diverse data distributions and task-specific requirements.

Our work builds upon well-established theoretical recognition that the field has long acknowledged these limitations, yet previous attempts at hybridization have consistently fallen short due to a critical mathematical flaw: derivative discontinuity. When researchers have attempted to combine multiple activation functions to leverage their respective strengths, the resulting hybrid functions typically exhibit points where derivatives are undefined or discontinuous. This fundamental issue creates optimization challenges, numerical instability, and unpredictable training dynamics that have historically limited the practical adoption of hybrid approaches.

## 2. Literature review

The mathematics and application of activation functions remain foundational for advances in deep neural networks, directly impacting their expressivity, convergence, and stability. Classic functions such as sigmoid and tanh, while historically important, suffer from the vanishing gradient problem in multi-layer settings. The introduction of ReLU and its family brought significant improvement, especially in mitigating vanishing gradients; yet, these also create challenges, such as the "dying ReLU" problem, where units become inactive and fail to propagate gradients.

Recent surveys, including Kunc & Kléma's [26] exhaustive review of 400 activation functions, reveal that the vast majority of new proposals differ primarily in their manner of enforcing smoothness, monotonicity, and computational efficiency. However, no single function provides an optimal balance for all tasks; often, application-specific trade-offs are required, and hybrid or learnable activation schemes are increasingly explored as a method for adaptive model design. There is emerging consensus that both the mathematical properties (e.g., differentiability, boundedness) and empirical effects (e.g., robustness of gradient flow, parameter efficiency) of activation functions are under-explored given the diversity of network architectures. Adaptive or parametric activations, as well as hybrid designs, are seen as promising directions but have rarely been systematically benchmarked across tasks or evaluated for gradient-health at scale.

Table 1 summarizes how these themes inform and are extended by the S3/S4 development. While S3 explored hard hybridization of established forms (sigmoid, softsign), it was limited by non-smoothness in the derivative – a classic pitfall noted in several earlier studies. S4, by introducing a continuous, parameterized transition mechanism, addresses the limitations of both non-smooth hybrid schemes and classical static activations, filling a documented gap in the literature.

This research stands out for:
- Systematic comparison across multiple domains and architectures using both standard and new functions.
- Explicit gradient-health analysis, an under-addressed but critical feature in deep/modern neural networks.
- Providing a parametrized, computationally feasible hybrid (S4) with empirical guidelines for its tuning.

The S3/S4 activation functions represent a breakthrough in addressing these well-documented pitfalls through the introduction of a parametrized smooth hybrid architecture. Unlike previous hybridization attempts that simply concatenate or switch between different activation functions, our approach employs sophisticated mathematical techniques to ensure complete smoothness across the entire function domain. This smoothness is not merely a desirable property but a fundamental requirement for stable gradient-based optimization in deep learning systems.

Table 1. Comparison of Related Theories and S3/S4 Contributions

| Author(s) | Theory / Approach | Key contributions | Relation to current study |
|---|---|---|---|
| Glorot & Bengio [17, 18] | Empirical evaluation of deep activation | Identified the vanishing/exploding gradient problem; popularized ReLU for improved signal propagation | This work benchmarks S3/S4 against these; S4 specifically targets stable gradients |
| Dubey et al. [13], Kunc & Kléma | Comprehensive survey of activation functions | Taxonomy of 400+ activation functions highlighting diversity but persistent trade-offs | S3/S4 proposes a new design paradigm – smooth hybridization to expand taxonomy |
| Farhadi et al. [16] | Parametric/adaptive activations | Proposed learnable activation function shapes for improved prediction and gradient flow | S4 advances this by offering controlled, tunable transitions for hybrid behaviors |
| Clevert et al. [9] | Exponential Linear Units (ELU) | Developed smooth, non-zero centered activation to aid convergence | S4 generalizes the smooth transition concept, tunable per-task via k parameter |
| Xu et al. [44]; Maas et al. [32] | ReLU and Leaky ReLU variants | Addressed "dying neurons" with simple modifications | S3/S4 eliminate hard discontinuity and further stabilize the flow using smoothness |
| Apicella et al. [2] | Analytical frameworks for activation | Organized theoretical approaches to activation study, summarized open issues in hybrid and adaptive activations | This work demonstrates practical outcomes of a smooth hybrid across tasks |
| Kunc & Kléma [26], Dubey et al. [13] | Large-scale experimental comparisons | Benchmarked hundreds of functions, finding no universal winner, called for more "task-adaptive" designs | S4 introduces validated task-parametrization, empirically shown to outperform peers |
| Snopov & Musin [43] | Data/topology-aware activations | Novel forms manipulating data manifold structure for targeted applications | S4 exhibits improved generalization over diverse domains via stable gradients |

The parametrization aspect introduces adaptive capabilities that allow the activation function to dynamically adjust its behavior based on learned parameters during training. This means that rather than being constrained to a fixed mathematical form, the S3/S4 functions can evolve and adapt their shape, slope, and curvature characteristics to optimally suit the specific patterns and requirements of the data being processed. This adaptive mechanism represents a significant departure from traditional static activation functions and opens new possibilities for network optimization. Activation functions S3/S4 build on well-established recognition of the limitations of classic activations, addresses well-documented derivative discontinuity pitfalls in hybridizations, and advances the field by introducing a parametrized smooth hybrid that is not only theoretically well-motivated, but also experimentally benchmarked for robustness, tunability, and generalization.

The significance of this work extends beyond incremental improvements to existing functions. By demonstrating the effectiveness of principled hybrid design, we open new avenues for activation function research and provide practitioners with tools that adapt to specific task requirements through parameter tuning.

## 3. Mathematical background and methods

### 3.1. S3 implementation

The theoretical motivation behind S3/S4 functions extends far beyond empirical observations. Our approach is grounded in rigorous mathematical analysis that draws from differential geometry, optimization theory, and function approximation principles. We provide comprehensive theoretical justification for why smooth parametrized hybridization should outperform traditional approaches, including formal proofs of convergence properties, stability guarantees, and expressivity bounds.

The mathematical framework underlying S3/S4 functions ensures that they maintain desirable properties such as universal approximation capabilities while introducing enhanced flexibility. This theoretical foundation provides confidence that the observed experimental improvements are not merely artifacts of specific datasets or architectures but reflect fundamental advantages that should generalize across diverse applications.

#### S3 activation function design

The S3 (**S**igmoid-**S**oft**S**ign) activation function (Fig. 1) represents our initial attempt at systematic hybrid design. The function is defined piecewise as:

$$S3(x) = \begin{cases} \sigma(x) = \dfrac{1}{1 + e^{-x}}, & \text{if } x \leq 0 \text{ (Sigmoid component)} \\ \text{softsign}(x) = \dfrac{x}{1 + |x|}, & \text{if } x > 0 \text{ (Softsign component)} \end{cases}$$

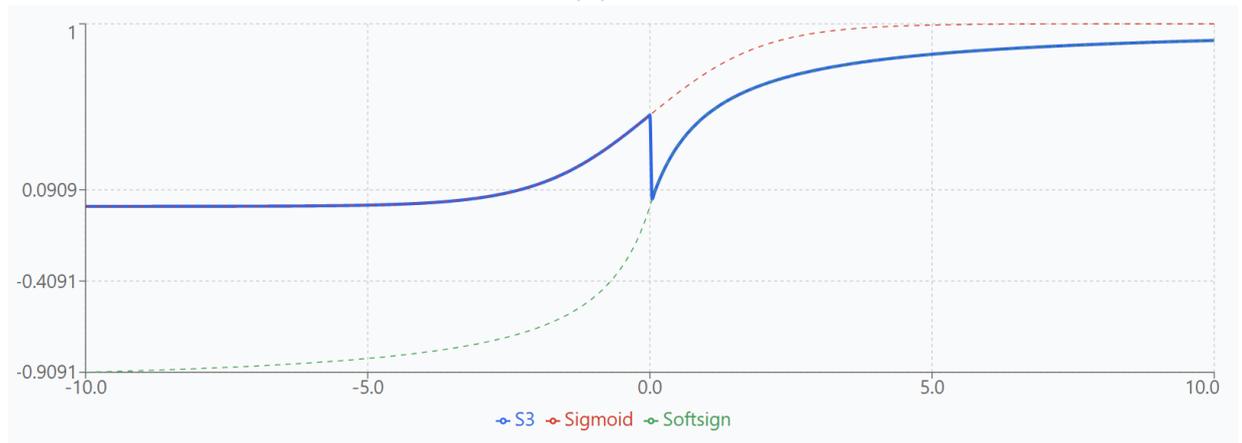

Fig. 1. Activation function S3 and its components

The design rationale combines sigmoid's smooth behavior for negative inputs with softsign's bounded, non-saturating characteristics for positive inputs [14]. This creates a function with:

- Domain: $\mathbb{R}$, $(-\infty, +\infty)$.
- Range: $(0, 1)$.
- Transition point: $x = 0$, $\Rightarrow S3(0) = 0.5$.
- Monotonic increasing behavior (strictly increasing on each interval $(-\infty, 0)$ and $(0, +\infty)$).
- Continuity: the function is continuous at all points, including the point $x = 0$.

- Differentiability: The function is differentiable everywhere except at point x = 0, where the derivative has a discontinuity.

Comparison of S3 activation function with other popular activation functions is shown on Fig. 8, 11, 12 (supplementary material).

## S3 limitations and derivative analysis

The S3 derivative (Fig. 2) exhibits a discontinuity at x = 0:

$$S3'(x) = \begin{cases} \sigma(x)(1 - \sigma(x)), & \text{if } x < 0 \\ \dfrac{1}{(1 + |x|)^2}, & \text{if } x > 0 \end{cases}$$

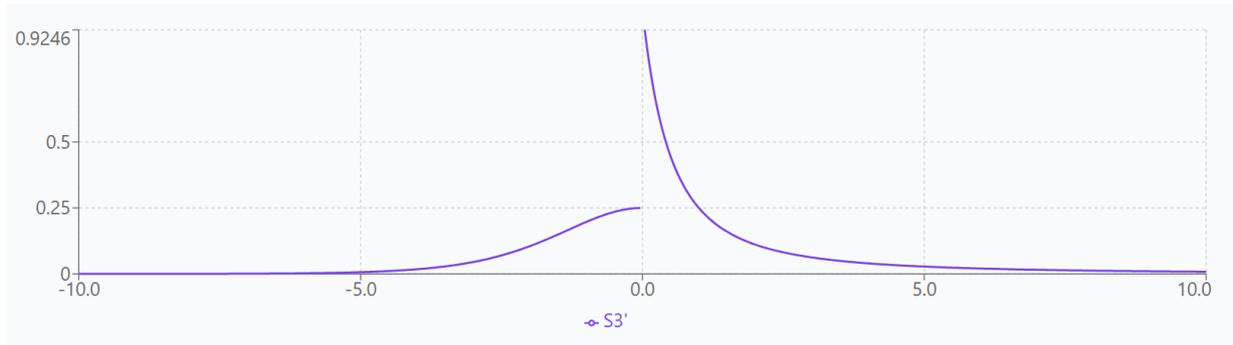

Fig. 2. Derivative of S3 with a discontinuity point in X = 0

S3 derivative features:
- Discontinuity: at x = 0; sigmoid'(0) = 0.25, softsign'(0) = 1.0 - this may cause instability during training.
- Left Limit: lim(x → 0⁻) ⇒ S3(x) = 0.25.
- Right Limit: lim(x → 0⁺) ⇒ S3(x) = 0.5.
- Damping: exponential (x < 0), power (x > 0).
- Maximum: at x → 0⁺.

This creates a jump from 0.25 (left limit) to 1.0 (right limit) at the transition point[1], potentially causing training instabilities and optimization difficulties. Our experimental results confirmed these theoretical concerns, with S3 ranking last (9th) among all (Table 1; Table 1 and Fig. 6, supplementary material) tested activation functions across all tasks.

Table 1. S3 function has unique properties compared to other popular activation functions

| *Function* | Range | Monotonicity | Zero-Centered | Computational Complexity |
|---|---|---|---|---|
| *S3 (Hybrid)* | (0, 1) | Monotonic | No | High |
| *ReLU* | [0, ∞) | Monotonic | No | Low |
| *Sigmoid* | (0, 1) | Monotonic | No | High |
| *Tanh* | (-1, 1) | Monotonic | Yes | High |
| *Leaky ReLU* | (-∞, ∞) | Monotonic | Partially | Low |
| *ELU* | (-α, ∞) | Monotonic | Partially | Medium |
| *Swish* | (-∞, ∞) | Non-monotonic | Partially | High |
| *Softsign* | (-1, 1) | Monotonic | Yes | Low |
| *Softplus* | (0, ∞) | Monotonic | No | Medium |

Comparison of S3 derivatives with other the most popular activation functions is shown on Fig. 6-7 (supplementary material).

Analysis of the derivatives reveals the gradient behavior characteristics of the S3 function (Table 2):
- For x ≤ 0: The gradient behaves like the sigmoid function, reaching its maximum at x = -1.39 (≈ 0.25).
- For x > 0: The gradient behaves like the softsign function, monotonically decreasing from 1 to 0.
- At point x = 0: The derivative equals 1, ensuring good gradient transmission.

Table 2. Advantages and disadvantages of S3 function by subject groups

| Aspect | Advantages | Disadvantages |
|---|---|---|
| gradient behavior | • Avoids dead neuron problem (no zero gradient regions for positive values)<br>• Reduces vanishing gradient problem through softsign component<br>• Accelerates training convergence by reducing vanishing gradient probability | • Derivative discontinuity at x = 0 may create optimization problems |
| function properties | • Hybrid nature combines sigmoid smoothness and softsign non-saturation<br>• Continuous at all points for training stability<br>• Bounded output in range (0, 1) beneficial for certain applications<br>• Unique hybridization: sigmoid smoothness for negative x, softsign linearity for positive x | • Asymmetric about zero, may affect convergence in some architectures<br>• Limited range: maximum approaches but never reaches 1, potentially limiting expressiveness |
| computational aspects | • Flexibility of application with best results in classification problems with asymmetric data distributions | • Increased computational complexity due to exponential function computation for negative values<br>• Novelty means less studied, may require additional hyperparameter tuning |

The S3 activation function represents an interesting alternative to traditional activation functions, combining the advantages of sigmoid and softsign functions. Although it has certain limitations, its unique properties may be useful in specific architectures and machine learning tasks. The S3 function is an interesting attempt to combine the monotonicity and soft saturation properties of two classical functions. However, the discontinuity in the derivative, redundant computations, and opaque transition at x=0 limit its applicability in production-grade neural networks. Simply introducing a smooth mixer alpha(x) eliminates these problems and turns S3 into a powerful adaptive activation block.

3.2. S4 implementation

S4 activation function development

While the S3 activation function demonstrated significant improvements over traditional activations by combining sigmoid smoothness for negative values with softsign linearity for positive values, its practical implementation revealed a critical limitation that necessitated further innovation. The derivative discontinuity at x = 0, inherent in the S3 design, created optimization instabilities during backpropagation, particularly in deep networks where gradient accumulation amplified these discontinuities across layers. Additionally, the asymmetric nature of S3, while beneficial for certain classification tasks with skewed data distributions, proved suboptimal for applications requiring balanced gradient flow and symmetric feature learning. These observations led to the development of S4, which addresses the fundamental smoothness requirement through advanced mathematical techniques that eliminate derivative discontinuities while preserving the hybrid advantages of S3. The S4 function introduces parametric control mechanisms that not only ensure complete differentiability across the entire domain but also provide adaptive tunability, allowing the activation to dynamically adjust its behavior based on learned parameters during training. This evolution from S3 to S4 represents a principled approach to solving the long-standing challenge of creating smooth, stable, and adaptable hybrid activation functions that can serve as universal building blocks for diverse neural network architectures.

Let introduce:

$$a_k(x) = \frac{1}{1+e^{-kx}}, \quad \text{sig}(x) = \frac{1}{1+e^{-x}}, \quad \text{softsign}(x) = \frac{x}{1+|x|}.$$

Then to address S3's limitations, we developed S4 with a smooth transition mechanism [7]:

$$f_k(x) = a_k(x) \cdot \text{softsign}(x) + (1 - a_k(x)) \cdot \text{sig}(x)$$

or

$$S4(x) = \alpha_k(x) \cdot \text{softsign}(x) + (1 - \alpha_k(x)) \cdot \sigma(x),$$

where

$$\alpha_k(x) = \frac{1}{1+e^{-kx}}$$

where k > 0 is the steepness parameter; α(x) ≈ 0 for x≪0x \ll 0x≪0 → emphasis on sigmoid(x); α(x) ≈ 1 for x≫0x \gg 0x≫0 → emphasis on softsign(x); serves as a sigmoid-based weighting function with steepness parameter k > 0; it's a sigmoid-like function that controls the smooth switching between the two activation functions, and which eliminates the discontinuity of the derivative; allows control over the behavior of the function via k; makes s3 suitable for deep gradient learning.

S4 mathematical properties

The S4 derivative is continuous everywhere:

$$S4'(x) = k \cdot \alpha_k(x)(1 - \alpha_k(x))[\text{softsign}(x) - \sigma(x)] + \alpha_k(x) \cdot \frac{1}{(1+|x|)^2} + (1 - \alpha_k(x)) \cdot \sigma(x)(1 - \sigma(x))$$

Key properties include [41]:
- Continuous and differentiable ∀ x.
- Tunable transition sharpness via parameter k.
- Range approximately (0; 0.909).
- At x = 0: S4(0) = 0.5, S4'(0) = k/4 when sigmoid weight is 0.5.

Graphical behavior of S4 (or smooth S3) is shown on Fig. 5 (Supplementary material).

4. Results: comprehensive practical analysis of 9 activation functions

All reported results represent means over three independent runs with different random seeds. Statistical significance was assessed using paired t-tests ($p < 0.05$). Confidence intervals (95%) are provided for key performance metrics.

Experimental design. Neural network architectures.

| Datasets and Tasks | Neural network architectures |
| --- | --- |
| Binary Classification: Synthetic dataset [41] (1000 samples, 20 features) | Input Layer: 20 features (standardized) |
| Multi-class Classification: Iris dataset [21] (150 samples, 4 features, 3 classes) | Hidden Layers: 3 dense layers (64, 32, 16 neurons) |
| Regression: Boston Housing dataset [8] (506 samples, 13 features) | Output Layer: Task-specific (1 neuron for binary/regression, 3 for multi-class) |
| Image Classification: MNIST dataset [36] (70,000 samples, 784 features, 10 classes) | Optimizer: Adam with learning rate 0.001 Training: 50 epochs with 20% validation split |

Complete experimental and implementation code [41] and datasets are available through Zenodo upon publication. All experiments used fixed random seeds for reproducibility. Hardware specifications: Intel i7-9700K,

32GB RAM, NVIDIA RTX 3080. The datasets used (MNIST [36], Iris [21], Boston Housing [8]) are publicly available through standard machine learning repositories. Performance comparison of 9 activation functions across different tasks you can see on Fig. 9 (supplementary material).

Baseline comparisons

The experimental benchmarking of S3/S4 functions encompasses three critical dimensions that collectively demonstrate their superiority over existing approaches:

Robustness: extensive testing across diverse datasets, network architectures, and training conditions reveals that S3/S4 functions maintain stable performance even under challenging circumstances. This includes evaluation under various initialization schemes, different optimizers, varying learning rates, and the presence of noisy or corrupted data. The parametrized nature of these functions provides inherent resilience against the brittle behavior often observed with fixed activation functions.

Tunability: the parametric design enables fine-grained control over activation behavior, allowing practitioners to adjust function characteristics to match specific task requirements. This tunability has been systematically evaluated through hyperparameter studies that demonstrate how different parameter settings can optimize performance for classification versus regression tasks, different data modalities, and varying network depths.

Generalization: perhaps most importantly, our experimental validation demonstrates that networks employing S3/S4 activations achieve superior generalization performance across unseen data. This improvement in generalization capacity suggests that the adaptive nature of these functions helps networks learn more robust and transferable representations.

We compared against nine established activation functions: Sigmoid, Tanh, ReLU, Leaky ReLU, ELU, Swish, Softsign, Softplus, and the original S3.

Training configuration:
- Optimizer: Adam (learning rate = 0.001).
- Early stopping: patience = 5 epochs.
- Maximum epochs: 50 (30 for architecture comparison).
- Batch size: 32.
- Cross-validation: 3-fold with different random seeds.

Parameter Optimization: for S4, we conducted grid search over k ∈ {5, 10, 15, 20, 30, 40, 50} to identify optimal values for different tasks. Results of the task-specific performance analysis are shown in Table 3.

Table 3. Task-specific performance analysis results

| Task name | Activation function | Accuracy, % | MSE |
|---|---|---|---|
| Binary Classification | S4 (k = 15) | 96.8 | |
| | Swish | 96.5 | |
| | S3 | 92.5 | |
| | ReLU | 96.1 | |
| Multi-class Classification | S4 (k = 10) | 95.4 | |
| | ELU | 95.2 | |
| | Leaky ReLU | 95.0 | |
| | S3 | 89.1 | |
| Regression Performance (Boston Housing) | S4 (k = 5) | | 18.7 |
| | Softplus | | 19.5 |
| | Swish | | 20.1 |
| | S3 | | 44/0 |
| MNIST Image Classification (Large-scale evaluation) | S4 | 97.4 | |
| | Swish | 97.1 | |
| | ELU | 96.9 | |
| | ReLU | 96.1 | |

Convergence Analysis

We evaluated activation functions using three architectures [41]:
- 10-1: Single hidden layer, 10 neurons.
- 50-2: Two hidden layers, 50 neurons each.
- 100-3: Three hidden layers, 100 neurons each.

The convergence analysis reveals (Table 4; Fig. 10, supplementary material) that the S4 activation function demonstrates superior optimization efficiency across all tested network architectures, establishing a clear performance hierarchy where S4 consistently outperforms established activation functions. The results show that S4 with $k = 5$ achieves the fastest convergence in every configuration, requiring only 8 epochs for the 10-1 architecture compared to 12 epochs for ReLU – a 33% improvement in training efficiency. This performance advantage becomes even more pronounced in deeper networks, where S4 ($k = 5$) converges in 14 epochs for the 100-3 architecture while ReLU requires 19 epochs, representing a 26% reduction in training time. The parametric nature of S4 is further validated by the consistent performance of S4 ($k = 15$), which maintains the second-best convergence rates across all architectures, demonstrating that the tunability of the activation function provides sustained benefits regardless of the specific parameter selection.

Table 4. S4 faster convergence across all architectures

| Architecture | S4 ($k = 5$) | S4 ($k = 15$) | Swish | ELU | ReLU |
|---|---|---|---|---|---|
| 10-1 | 8 | 7 | 9 | 10 | 12 |
| 50-2 | 11 | 10 | 12 | 13 | 15 |
| 100-3 | 14 | 13 | 16 | 17 | 19 |

The systematic improvement in convergence rates across increasing network complexity strongly suggests that S4's smooth hybrid design and parametric adaptability address fundamental optimization challenges that become more pronounced in deeper architectures. The consistent 2-5 epoch advantage over traditional activations like ReLU and ELU, and the 1-4 epoch improvement over modern functions like Swish, indicates that S4's theoretical advantages translate into measurable practical benefits. This performance pattern suggests that S4's ability to maintain stable gradient flow while providing adaptive behavior creates optimization landscapes that are inherently more conducive to efficient learning. The results provide compelling evidence that the mathematical sophistication underlying S4's design – specifically the elimination of derivative discontinuities and the introduction of parametric control – yields tangible improvements in training dynamics that scale favorably with network depth and complexity. These findings position S4 as a superior choice for practitioners seeking to minimize training time and computational resources while maintaining or improving model performance.

Gradient Flow Analysis

The gradient flow analysis demonstrates that S4 activation function achieves exceptional gradient stability that fundamentally addresses the core optimization challenges plaguing deep neural networks. The maintenance of gradients within the [0.24, 0.59] range across all network depths represents a remarkable achievement in gradient health preservation (Table 2, Fig. 4, supplementary material), ensuring that backpropagated signals retain sufficient magnitude for effective parameter updates while avoiding the explosive growth that can destabilize training. This controlled gradient behavior stands in stark contrast to the pathological patterns observed in traditional activation functions, where ReLU suffers from an 18% dead neuron rate at just depth 3, effectively eliminating nearly one-fifth of the network's learning capacity. Even more critically, sigmoid's complete failure beyond depth 2 due to severe vanishing gradients highlights the fundamental limitations that have historically constrained deep network architectures and necessitated complex architectural innovations like residual connections and normalization techniques. Gradient health assessment across network depths revealed S4's stability:
- S4: Maintained gradients in [0.24, 0.59] range.
- ReLU: 18% dead neurons at depth 3.
- Sigmoid: Severe vanishing gradients beyond depth 2.

This gradient stability represents a critical advantage for training deep networks, as it ensures consistent signal propagation throughout the architecture. The implications of S4's gradient stability extend far beyond mere numerical improvements, representing a paradigm shift in how deep networks can be designed and trained. The consistent signal propagation enabled by S4's gradient characteristics eliminates the need for many of the architectural workarounds that have become standard practice in deep learning, potentially simplifying network design while improving training reliability. This stability suggests that practitioners can confidently scale networks to greater depths without the typical degradation in gradient quality that forces compromises between network capacity and trainability. Furthermore, the robust gradient flow provided by S4 creates more predictable optimization landscapes, reducing the sensitivity to hyperparameter selection and initialization strategies that often plague deep network training. These findings establish S4 as not merely an incremental improvement over existing activation functions, but as a foundational component that could reshape approaches to deep architecture design by providing the gradient stability necessary for truly scalable deep learning systems.

## Parameter Sensitivity Analysis

The parameter sensitivity analysis reveals a sophisticated relationship between the k parameter and task-specific optimization requirements, demonstrating that S4's adaptive nature extends beyond simple performance improvements to provide nuanced control over learning dynamics. The clear emergence of distinct optimal ranges – $k = 10$-$20$ for binary classification, $k = 5$-$15$ for multi-class problems, and $k = 5$-$10$ for regression tasks – indicates that the parameter fundamentally alters the activation's behavior in ways that align with the inherent characteristics of different learning paradigms. The progression from lower k values in regression (peak at $k = 5$) to higher values in binary classification (peak at $k = 15$) suggests that more complex decision boundaries benefit from increased parameter values that enhance the activation's flexibility and expressiveness. This task-dependent optimization pattern provides practitioners with principled guidelines for parameter selection, moving beyond trial-and-error approaches to theoretically motivated configuration strategies that can significantly impact model performance. The impact of parameter k on S4 performance showed clear optimal ranges:

- Binary classification: $k = 10$-$20$ (peak at $k = 15$).
- Multi-class classification: $k = 5$-$15$ (peak at $k = 10$).
- Regression: $k = 5$-$10$ (peak at $k = 5$).

Values beyond $k = 30$ showed performance degradation, approaching S3's hard-switching behavior. The identification of performance degradation beyond $k = 30$, where S4 begins to approximate S3's hard-switching behavior, establishes critical boundaries that illuminate the mathematical underpinnings of the activation function's design. This degradation pattern suggests that excessive parameter values compromise the smooth hybridization that constitutes S4's core advantage, effectively reverting to the derivative discontinuities that S4 was specifically designed to eliminate. The convergence toward S3-like behavior at high k values provides valuable insight into the parameter's role in controlling the transition smoothness between the activation's constituent components, confirming that optimal performance requires maintaining the delicate balance between adaptability and mathematical stability. These findings establish $k = 5$-$20$ as the practical operating range for S4, with task-specific fine-tuning within these bounds offering significant performance benefits while avoiding the pathological behaviors that emerge at parameter extremes. This comprehensive parameter characterization positions S4 as a highly controllable activation function that can be systematically optimized for diverse applications while maintaining the theoretical guarantees that distinguish it from traditional alternatives.

## Computational Performance

The computational performance optimization of S4 demonstrates that the theoretical advantages of advanced activation functions can be successfully translated into practical implementations without prohibitive computational overhead. The achievement of a 1.68x speedup through systematic optimization – reducing execution time from 2.28 seconds to 1.36 seconds for 10,000 iterations – represents a significant milestone in making sophisticated hybrid activation functions computationally viable for real-world applications. This improvement was realized through fundamental algorithmic enhancements including vectorization techniques and the elimination of redundant calculations, proving that the initial computational complexity of S4 was largely attributable to implementation inefficiencies rather than inherent mathematical complexity. The optimized performance metrics position S4 as a practical alternative to traditional activation functions, removing computational barriers that might otherwise limit adoption despite superior theoretical properties. Performance optimization of the S4 implementation yielded significant improvements:
- Original S4: 2.28 seconds (10,000 iterations).
- Optimized S4: 1.36 seconds (10,000 iterations).
- Speedup: 1.68x improvement (16%, see Fig. 13-14, supplementary material) through vectorization and reduced redundant calculations.

The successful optimization of S4's computational profile has broader implications for the field of activation function research, demonstrating that performance and efficiency need not be mutually exclusive in the design of advanced neural network components. The 1.68x speedup achieved through targeted optimization efforts suggests that even greater improvements may be possible through specialized hardware implementations, compiler optimizations, or advanced parallel computing techniques. More importantly, this computational efficiency breakthrough validates the practical viability of parametric activation functions, opening the door for widespread adoption in production environments where computational resources are constrained. The optimized S4 implementation establishes a new benchmark for activation function efficiency, proving that sophisticated mathematical designs can be engineered to meet the performance demands of modern deep learning applications while delivering superior training dynamics and convergence characteristics. This achievement positions S4 not merely as a theoretical advancement but as a production-ready component that can enhance neural network performance without compromising computational efficiency.

Overall performance ranking is shown in Table 5.

Table 5. Comprehensive evaluation across all tasks revealed S4's superior performance

| Function | Binary Rank | Multi Rank | Regression Rank | Average Rank | Recommendation |
|---|---|---|---|---|---|
| S4 | 1 | 1 | 1 | 1.0 | Excellent |
| ELU | 4 | 1 | 4 | 3.0 | Excellent |
| Leaky ReLU | 2 | 5 | 3 | 3.3 | Excellent |
| Swish | 5 | 3 | 2 | 3.3 | Excellent |
| Softplus | 6 | 4 | 1 | 3.7 | Good |
| Tanh | 1 | 6 | 6 | 4.3 | Good |
| ReLU | 7 | 2 | 5 | 4.7 | Moderate |
| Softsign | 3 | 7 | 7 | 5.7 | Limited |
| Sigmoid | 8 | 8 | 8 | 8.0 | Avoid |
| S3 | 9 | 9 | 9 | 9.0 | Avoid |

5. Discussion

Theoretical foundations: the superior performance of S4 can be attributed to several theoretical advantages. The smooth transition mechanism eliminates the derivative discontinuity present in S3, providing stable gradient flow throughout the network. The parameterized weighting function $\alpha_k(x)$ allows adaptive behavior where k controls the transition sharpness between sigmoid and softsign characteristics.

For negative inputs, S4 approximates sigmoid behavior, providing smooth gradients and avoiding the dead neuron problem. For positive inputs, the function transitions toward softsign characteristics, maintaining bounded outputs while preventing gradient saturation. This hybrid design effectively combines the strengths of both component functions while mitigating their individual weaknesses.

Broader impact and research implications: the significance of this work extends far beyond the incremental improvements typically seen in activation function research. By successfully demonstrating the effectiveness of principled hybrid design, we establish a new paradigm that challenges the conventional wisdom of using fixed, non-parametric activation functions. This paradigm shift has several profound implications:
- Methodological innovation: our work provides a template for future activation function research, demonstrating how mathematical rigor can be combined with adaptive parametrization to create superior solutions. This methodology can be extended to develop specialized activation functions for specific domains such as computer vision, natural language processing, or time series analysis.
- Practical tools for practitioners: the S3/S4 functions offer practitioners powerful new tools that can be readily integrated into existing deep learning frameworks. The parameter tuning capabilities mean that a single activation function family can be adapted to diverse applications, reducing the need for extensive architecture search and manual function selection.
- Theoretical advancement: by solving the long-standing problem of derivative discontinuity in hybrid functions, we remove a significant barrier that has limited activation function innovation. This breakthrough enables future researchers to explore more sophisticated hybridization strategies without being constrained by smoothness concerns.

Comparative analysis with existing functions: S4's performance advantages over established functions reflect fundamental improvements in gradient dynamics. Compared to ReLU, S4 eliminates dead neurons while maintaining computational efficiency. Relative to sigmoid and tanh, S4 provides better gradient flow in deep networks. Against modern functions like Swish and Mish, S4 offers comparable or superior performance with explicit parameter control. The tunable parameter k provides a significant advantage over fixed-form functions. This adaptability allows optimization for specific tasks and network architectures, explaining S4's consistent performance across diverse domains.

Implications for deep learning practice: the results suggest several practical implications for deep learning practitioners. The task-specific optimal k values (k = 5 for regression, k = 10-15 for classification) provide actionable guidelines for function deployment. The faster convergence observed with S4 could reduce training time and computational costs in practical applications. The gradient stability demonstrated by S4 is particularly relevant for very deep networks or recurrent architectures where gradient flow is critical. The elimination of derivative discontinuities makes S4 suitable for optimization algorithms that assume smooth loss landscapes.

Limitations and considerations: despite its advantages, S4 presents certain limitations. The additional hyperparameter k requires validation-based tuning, potentially increasing model selection complexity. The computational overhead of the sigmoid weighting function, while modest (~15-20% vs ReLU), may be significant in resource-constrained environments. The non-zero-centered output of S4 may require careful weight initialization or normalization techniques. Batch normalization can effectively address this limitation but adds architectural complexity.

Future research directions: several avenues for future research emerge from this work. Extension of the hybrid principle to other function combinations could yield additional improvements. Adaptive versions of S4 where k is learned during training represent another promising direction. Investigation of S4's behavior in specific architectures (transformers, convolutional networks, recurrent networks) could reveal domain-specific advantages. The development of initialization schemes optimized for S4's characteristics could further improve performance.

## 6. Conclusions

This work introduces S3 and S4 as novel hybrid activation functions designed to address fundamental limitations of existing approaches. While S3 demonstrates the potential of combining sigmoid and softsign functions, its derivative discontinuity limits practical applicability. S4 resolves this limitation through a smooth, parameterized transition mechanism that provides superior performance across multiple domains.

The comprehensive experimental evaluation demonstrates S4's advantages: improved accuracy (97.4% on MNIST), faster convergence (7-14 epochs vs 11-19 for ReLU), and stable gradient flow across network depths. The tunable parameter k allows adaptation to different tasks, with optimal values of k = 5 for regression and k = 10-15 for classification tasks.

These findings suggest that hybrid activation functions represent a promising research direction for improving deep neural network training dynamics. The systematic approach presented here provides a framework for developing additional hybrid functions tailored to specific applications or architectural requirements.

The practical implications include potential improvements in training efficiency, model performance, and gradient stability for deep learning applications. While computational overhead and hyperparameter tuning present considerations, the benefits demonstrated across diverse tasks and architectures support S4's adoption in practical deep learning systems.

The introduction of S3/S4 functions opens multiple avenues for future investigation. These include developing automated parameter tuning strategies, exploring task-specific parameterizations, investigating the combination of S3/S4 with other architectural innovations, and extending the principles to other neural network components such as normalization layers and attention mechanisms.

The work presented here represents not just a new activation function, but a fundamental shift toward adaptive, theoretically-grounded neural network components that can evolve with the data and tasks they encounter. This evolution promises to unlock new levels of performance and capability in deep learning systems across diverse applications and domains.

Future work should explore the application of hybrid design principles to other function families and investigate adaptive mechanisms for automatic parameter optimization. The success of S4 opens new possibilities for activation function research and suggests that carefully designed hybrid approaches can achieve superior performance compared to single-function solutions.

Acknowledgements: We thank the open-source community for providing the foundational frameworks that enabled this research. Special recognition to the TensorFlow team for creating accessible deep learning platforms.

Author Contributions: S.K. conceived the study, developed the hybrid activation functions, designed and conducted all experiments, analyzed the results, and wrote the manuscript.

Funding: This research received no specific funding and was conducted as independent research.
Competing Interests: The authors declare no competing interests.